# A HYBRID REASONING MODEL FOR INDIRECT ANSWERS

**Nancy Green**
Department of Computer Science
University of Delaware
Newark, DE 19716, USA
Internet: green@udel.edu

**Sandra Carberry**
Department of Computer Science
University of Delaware
Visitor: Inst. for Research in Cognitive Science
University of Pennsylvania
Internet: carberry@udel.edu

## Abstract

This paper presents our implemented computational model for interpreting and generating indirect answers to Yes-No questions. Its main features are 1) a discourse-plan-based approach to implicature, 2) a reversible architecture for generation and interpretation, 3) a hybrid reasoning model that employs both plan inference and logical inference, and 4) use of stimulus conditions to model a speaker's motivation for providing appropriate, unrequested information. The model handles a wider range of types of indirect answers than previous computational models and has several significant advantages.

## 1. INTRODUCTION

Imagine a discourse context for (1) in which R's use of just (1d) is intended to convey a No, i.e., that R is *not* going shopping tonight. (By convention, square brackets indicate that the enclosed text was not explicitly stated.) The part of R's response consisting of (1d) - (1e) is what we call an *indirect answer* to a Yes-No question, and if (1c) had been uttered, (1c) would have been called a *direct answer*.

```
1.a. Q: I need a ride to the mall.
  b.    Are you going shopping tonight?
  c. R: [no]
  d.    My car's not running.
  e.    The rear axle is broken.
```

According to one study of spoken English [Stenström, 1984], 13 percent of responses to Yes-No questions were indirect answers. Thus, the ability to interpret indirect answers is required for robust dialogue systems. Furthermore, there are good reasons for generating indirect answers instead of just *yes*, *no*, or *I don't know*. First, they may provide information which is needed to avoid misleading the questioner [Hirschberg, 1985]. Second, they contribute to an efficient dialogue by anticipating follow-up questions. Third, they may be used for social reasons, as in (1).

This paper provides a computational model for the interpretation and generation of indirect answers to Yes-No questions in English. More precisely, by a *Yes-No question* we mean one or more utterances used as a request by Q (the questioner) that R (the responder) convey R's evaluation of the truth of a proposition $p$. An indirect answer implicitly conveys via one or more utterances R's evaluation of the truth of the questioned proposition $p$, i.e. that $p$ is true, that $p$ is false, that there is some truth to $p$, that $p$ may be true, or that $p$ may be false. Our model presupposes that Q's question has been understood by R as intended by Q, that Q's request was appropriate, and that Q and R are engaged in a cooperative goal-directed dialogue. The interpretation and generation components of the model have been implemented in Common Lisp on a Sun SPARCstation.

The model employs an agent's pragmatic knowledge of how language typically is used to answer Yes-No questions in English to constrain the process of generating and interpreting indirect answers. This knowledge is encoded as a set of domain-independent discourse plan operators and a set of coherence rules, described in section 2.[1] Figure 1 shows the architecture of our system. It is reversible in that the same pragmatic knowledge is used by the interpretation and generation modules. The interpretation algorithm, described in section 3, is a hybrid approach employing both plan inference and logical inference to infer R's discourse plan. The generation algorithm, described in section 4, constructs R's discourse plan in two phases. During the first phase, *stimulus conditions* are used to trigger goals to include appropriate, extra information in the response plan. In the second phase, the response plan is pruned to eliminate parts which can be inferred by Q.

---

[1]Our main sources of data were previous studies [Hirschberg, 1985, Stenström, 1984], transcripts of naturally occurring two-person dialogue [American Express transcripts, 1992], and constructed examples.

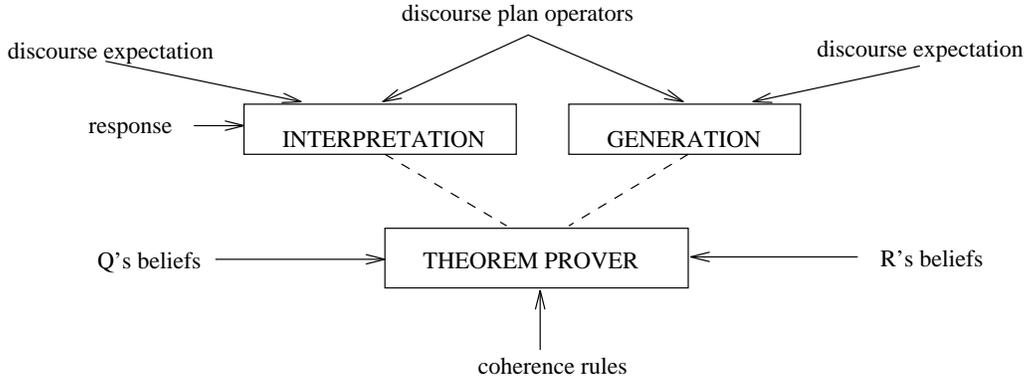

Figure 1: Architecture of system

## 2. PRAGMATIC KNOWLEDGE

Linguists (e.g. see discussion in [Levinson, 1983]) have claimed that use of an utterance in a dialogue may create shared expectations about subsequent utterances. In particular, a Yes-No question creates the discourse expectation that R will provide R's evaluation of the truth of the questioned proposition $p$. Furthermore, Q's assumption that R's response is relevant triggers Q's attempt to interpret R's response as providing the requested information. We have observed that coherence relations similar to the subject-matter relations of Rhetorical Structure Theory (RST) [Mann and Thompson, 1987] can be used in defining constraints on the relevance of an indirect answer. For example, the relation between the (implicit) direct answer in (2b) and each of the indirect answers in (2c) - (2e) is similar to RST's relations of Condition, Elaboration, and (Volitional) Cause, respectively.

2.a. Q: Are you going shopping tonight?
  b. R: [yes]
  c.    if I finish my homework
  d.    I'm going to Macy's
  e.    Winter clothes are on sale

Furthermore, for Q to interpret any of (2c) - (2e) as conveying an affirmative answer, Q must believe that R intended Q to recognize the *relational proposition* holding between the indirect answer and (2b), e.g. that (2d) is an elaboration of (2b). Also, coherence relations hold between parts of an indirect answer consisting of multiple utterances. For example, (1e) describes the cause of the failure reported in (1d). Finally, we have observed that different relations are usually associated with different types of answers. Thus, a speaker who has inferred a plausible coherence relation holding between an indirect answer and a possible (implicit) direct answer may be able to infer the direct answer. (If more than one coherence relation

```
( (Plausible (cr-obstacle
    ((not (in-state ?stateq ?tq))
     (not (occur ?eventp ?tp)))))
<-
(state ?stateq)
(event ?eventp)
(timeperiod ?tq)
(timeperiod ?tp)
(before ?tq ?tp)
(app-cond ?stateq ?eventp)
(unless (in-state ?stateq ?tq))
(unless (occur ?eventp ?tp)))
```

Figure 2: A coherence rule for *cr-obstacle*

is plausible, or if the same coherence relation is used with more than one type of answer, then the indirect answer may be ambiguous.)

In our model we formally represent the coherence relations which constrain indirect answers by means of *coherence rules*. Each rule consists of a consequent of the form *(Plausible (CR q p))* and an antecedent which is a conjunction of conditions, where *CR* is the name of a coherence relation and $q$ and $p$ are formulae, symbols prefixed with "?" are variables, and all variables are implicitly universally quantified. Each antecedent condition represents a condition which is true iff it is believed by R to be mutually believed with Q.[2] Each rule represents sufficient conditions for the plausibility of *(CR q p)* for some *CR*, $q$, $p$. An example of one of the rules describing the Obsta-

---

[2] Our model of R's beliefs (and similarly for Q's), represented as a set of Horn clauses, includes 1) general world knowledge presumably shared with Q, 2) knowledge about the preceding discourse, and 3) R's beliefs (including "weak beliefs") about Q's beliefs. Much of the shared world knowledge needed to evaluate the coherence rules consists of knowledge from domain plan operators.

```
(Answer-yes s h ?p):                    (Answer-no s h ?p):
  Applicability conditions:               Applicability conditions:
    (discourse-expectation                  (discourse-expectation
        (informif s h ?p))                      (informif s h ?p))
    (believe s ?p)                          (believe s (not ?p))
  Nucleus:                                Nucleus:
    (inform s h ?p)                         (inform s h (not ?p))
  Satellites:                             Satellites:
    (Use-condition s h ?p)                  (Use-otherwise s h (not ?p))
    (Use-cause s h ?p)                      (Use-obstacle s h (not ?p))
    (Use-elaboration s h ?p)                (Use-contrast s h (not ?p))
  Primary goals:                          Primary goals:
    (BMB h s ?p)                            (BMB h s (not ?p))
```

Figure 3: Discourse plan operators for Yes and No answers

cle relation[3] is shown in Figure 2. The predicates used in the rule are defined as follows: *(in-state p t)* denotes that $p$ holds during $t$, *(occur p t)* denotes that $p$ happens during $t$, *(state x)* denotes that the type of $x$ is state, *(event x)* denotes that the type of $x$ is event, *(timeperiod t)* denotes that $t$ is a time interval, *(before t1 t2)* denotes that $t1$ begins before or at the same time as $t2$, *(app-cond q p)* denotes that $q$ is a plausible enabling condition for doing $p$, and *(unless p)* denotes that $p$ is not provable from the beliefs of the reasoner. For example, this rule describes the relation between (1d) and (1c), where (1d) is interpreted as *(not (in-state (running R-car) Present))* and (1c) as *(not (occur (go-shopping R) Future))*. That is, this relation would be plausible if Q and R share the belief that a plausible enabling condition of a subaction of a plan for R to go shopping at the mall is that R's car be in running condition.

In her study of responses to questions, Stenström [Stenström, 1984] found that direct answers are often accompanied by extra, relevant information,[4] and noted that often this extra information is similar in content to an indirect answer. Thus, the above constraints on the relevance of an indirect answer can serve also as constraints on information accompanying a direct answer. For maximum generality, therefore, we went beyond our original goal of handling indirect answers to the goal of handling what we call full answers. A full answer consists of an implicit or explicit direct answer (which we call the *nucleus*) and, possibly, extra, relevant information (*satellites*).[5] In our model, we represent each type of full answer as a (top-level) discourse plan operator. By representing answer types as plan operators, generation can be modeled as plan construction, and interpretation as plan recognition.

Examples of (top-level) operators describing a full Yes answer and a full No answer are shown in Figure 3.[6] To explain our notation, $s$ and $h$ are constants denoting speaker (R) and hearer (Q), respectively. Symbols prefixed with "?" denote propositional variables. The variables in the header of each top-level operator will be instantiated with the questioned proposition. In interpreting example (1), ?p would be instantiated with the proposition that R is going shopping tonight. Thus, instantiating the Answer-No operator in Figure 3 with this proposition would produce a plan for answering that R is *not* going shopping tonight. Applicability conditions are necessary conditions for appropriate use of a plan operator. For example, it is inappropriate for R to give an affirmative answer that $p$ if R believes $p$ is false. Also, an answer to a Yes-No question is not appropriate unless $s$ and $h$ share the discourse expectation that $s$ will provide $s$'s evaluation of the truth of the questioned proposition $p$, which we denote as *(discourse-expectation (informif s h p))*. Primary goals describe the intended effects of the plan operator. We use *(BMB h s p)* to denote that $h$ believes it mutually believed with $s$ that $p$ [Clark and Marshall, 1981].

In general, the nucleus and satellites of a discourse plan operator describe primitive or non-primitive communicative acts. Our formalism al-

---

[3]While Obstacle is not one of the original relations of RST, it is similar to the causal relations of RST.

[4]61 percent of direct No answers and 24 percent of direct Yes answers

[5]The terms *nucleus* and *satellite* have been borrowed from RST to reflect the informational constraints within a full answer. Note that according to RST, a property of the nucleus is that its removal results in incoherence. However, in our model, a direct answer may be removed without causing incoherence, provided that it is inferable from the rest of the response.

[6]The other top-level operators in our model, *Answer-hedged*, *Answer-maybe*, and *Answer-maybe-not*, represent the other answer types handled.

```
(Use-obstacle s h ?p):
;; s tells h of an obstacle explaining
;; the failure ?p
  Existential variable: ?q
  Applicability conditions:
    (believe s (cr-obstacle ?q ?p))
    (Plausible (cr-obstacle ?q ?p))
  Stimulus conditions:
    (explanation-indicated s h ?p ?q)
    (excuse-indicated s h ?p ?q)
  Nucleus:
    (inform s h ?q)
  Satellites:
    (Use-elaboration s h ?q)
    (Use-obstacle s h ?q)
    (Use-cause s h ?q)
  Primary goals:
    (BMB h s (cr-obstacle ?q ?p))
```

Figure 4: Discourse plan operator for Obstacle

lows zero, one, or more occurrences of a satellite in a full answer, and the expected (but not required) order of nucleus and satellites is the order they are listed in the operator. *(inform s h p)* denotes the primitive act of s informing h that p. The satellites in Figure 3 refer to non-primitive acts, described by discourse plan operators which we have defined (one for each coherence relation used in a full answer). For example, *Use-obstacle*, a satellite of *Answer-no* in Figure 3, is defined in Figure 4.

To explain the additional notation in Figure 4, *(cr-obstacle q p)* denotes that the coherence relation named obstacle holds between q and p. Thus, the first applicability condition can be glossed as requiring that s believe that the coherence relation holds. In the second applicability condition, *(Plausible (cr-obstacle q p))* denotes that, given what s believes to be mutually believed with h, the coherence relation *(cr-obstacle q p)* is plausible. This sort of applicability condition is evaluated using the coherence rules described above.

Stimulus conditions describe conditions motivating a speaker to include a satellite during plan construction. They can be thought of as triggers which give rise to new speaker goals. In order for a satellite to be selected during generation, all of its applicability conditions and at least one of its stimulus conditions must hold. While stimulus conditions may be derivative of principles of cooperativity [Grice, 1975] or politeness [Brown and Levinson, 1978], they provide a level of precompiled knowledge which reduces the amount of reasoning required for content-planning. For example, Figure 5 depicts the discourse plan which would be constructed by R (and must be inferred by Q) for (1). The first stimulus condition of *Use-obstacle*, which is defined as holding whenever s suspects that h would be surprised that p holds, describes R's reason for including (1e). The second stimulus condition, which is defined as holding whenever s suspects that the Yes-No question is a prerequest [Levinson, 1983], describes R's reason for including (1d).[7]

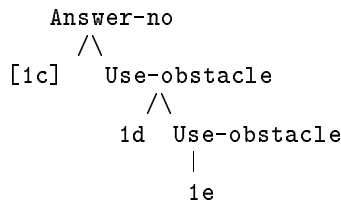

Figure 5: Discourse plan underlying (1d) - (1e)

## 3. INTERPRETATION

We assume that interpretation of dialogue is controlled by a Discourse Model Processor (DMP), which maintains a Discourse Model [Carberry, 1990] representing what Q believes R has inferred so far concerning Q's plans. The discourse expectation generated by a Yes-No question leads the DMP to invoke the answer recognition process to be described in this section. If answer recognition is unsuccessful, the DMP would invoke other types of recognizers for handling less preferred types of responses, such as *I don't know* or a clarification subdialogue. To give an example of where our recognition algorithm fits into the above framework, consider (4).

```
4a. Q: Is Dr. Smith teaching CS1 next fall?
 b. R: Do you mean Dr. Smithson?
 c. Q: Yes.
 d. R: [no]
 e.    He will be on sabbatical next fall.
 f.    Why do you ask?
```

Note that a request for clarification and its answer are given in (4b) - (4c). Our recognition algorithm, when invoked with (4e) - (4f) as input, would infer an *Answer-no* plan accounting for (4e) and satisfying the discourse expectation generated by (4a).

When invoked by the DMP, our interpretation module plays the role of the questioner Q. The inputs to interpretation in our model consist of

---
[7]Stimulus conditions are formally defined by rules encoded in the same formalism as used for our coherence rules. A full description of the stimulus conditions used in our model can be found in [Green, in preparation].

1) the set of discourse plan operators and the set of coherence rules described in section 2, 2) Q's beliefs, 3) the discourse expectation *(discourse-expectation (informif s h p))*, and 4) the semantic representation of the sequence of utterances performed by R during R's turn. The output is a partially ordered set (possibly empty) of answer discourse plans which it is plausible to ascribe to R as underlying R's response. The set is ordered by plausibility using preference criteria. Note that we assume that the final choice of a discourse plan to ascribe to R is made by the DMP, since the DMP must select an interpretation consistent with the interpretation of any remaining parts of R's turn not accounted for by the answer discourse plan, e.g. (4f).

To give a high-level description of our answer interpretation algorithm, first, each (top-level) answer discourse plan operator is instantiated with the questioned proposition from the discourse expectation. For example (1), each answer operator would be instantiated with the proposition that R is going shopping tonight. Next, the answer interpreter must verify that the applicability conditions and primary goals which would be held by R if R were pursuing the plan are consistent with Q's beliefs about R's beliefs and goals. Consistency checking is implemented using a Horn clause theorem-prover. For all candidate answer plans which have not been eliminated during consistency checking, recognition continues by attempting to match the utterances in R's turn to the actions specified in the candidates. However, no candidate plan may be constructed which violates the following structural constraint. Viewing a candidate plan's structure as a tree whose leaves are primitive acts from which the plan was inferred, no subtree $T_i$ may contain an act whose sequential position in the response is included in the range of sequential positions in the response of acts in a subtree $T_j$ having the same parent node as $T_i$. For example, (5e) cannot be interpreted as related to (5c) by *cr-obstacle*, due to the occurrence of (5d) between (5c) and (5e). Note that a more coherent response would consist of the sequence, (5c), (5e), (5d).

```
5.a. Q: Are you going shopping tonight?
  b. R: [no]
  c.    My car's not running.
  d.    Besides, I'm too tired.
  e.    The timing belt is broken.
```

To recognize a subplan for a non-primitive action, e.g. *Use-obstacle* in Figure 4, a similar procedure is used. Note that any applicability condition of the form *(Plausible (CR q p))* is defined to be consistent with Q's beliefs if it is provable, i.e., if the antecedents of a coherence rule for *CR* are true with respect to what Q believes to be mutually believed with R. The recognition process for non-primitive actions differs in that these operators contain existential variables which must be instantiated. In our model, the answer interpreter first attempts to instantiate an existential variable with a proposition from R's response. For example (1), the existential variable $?q$ of *Use-obstacle* would be instantiated with the proposition that R's car is not running. However, if (1d) was not explicitly stated by R, i.e., if R's response had just consisted of (1e), it would be necessary for $?q$ to be instantiated with a *hypothesized* proposition, corresponding to (1d), to understand how (1e) relates to R's answer. The answer interpreter finds the hypothesized proposition by a subprocedure we refer to as *hypothesis generation*.

Hypothesis generation is constrained by the assumption that R's response is coherent, i.e., that (1e) may play the role of a satellite in a subplan of some Answer plan. Thus, the coherence rules are used as a source of knowledge for generating hypotheses. Hypothesis generation begins with initializing the root of a tree of hypotheses with a proposition $p_0$ to be related to a plan, e.g. the proposition conveyed by (1e). A tree of hypotheses is constructed by expanding each of its nodes in breadth-first order until all goal nodes (as defined below) have been reached, subject to a limit on the depth of the breadth-first search.[8] A node containing a proposition $p_i$ is expanded by searching for all propositions $p_{i+1}$ such that for some coherence relation $CR$ which may be used in the type of answer being recognized, *(Plausible (CR $p_i$ $p_{i+1}$))* holds from Q's point of view. (The search is implemented using a Horn clause theorem prover.)

The plan operator invoking hypothesis generation has a partially instantiated applicability condition of the form, *(Plausible (CR ?q p))*, where $CR$ is a coherence relation, $p$ is the proposition that was used to instantiate the header variable of the operator, and $?q$ is the operator's existential variable. Since the purpose of the search is to find a proposition $q$ with which to instantiate $?q$, a goal node is defined as a node containing a proposition $q$ satisfying the above condition. (E.g. in Figure 6 $p_0$ is the proposition conveyed by (1e), $p_1$ is the proposition conveyed by (1d), $p_0$ and $p_1$ are plausibly related by *cr-obstacle*, $p_2$ is the proposition conveyed by a No answer to (1a), $p_1$ and $p_2$ are plausibly related by *cr-obstacle*, $p_2$ is a goal node, and therefore, $p_1$ will be used to instantiate the existential variable $?q$ in *Use-obstacle*.)

After the existential variable is instantiated, plan recognition proceeds as described above at the point where the remaining conditions are checked for consistency.[9] For example, as recognition of the *Use-obstacle* subplan proceeds, (1e) would be recognized as the realization of a *Use-obstacle* satellite of this *Use-obstacle* subplan. Ultimately, the inferred plan would be the same as that shown in Figure 5, except that (1d) would be marked as hypothesized.

---

[8] Placing a limit on the maximum depth of the tree is reasonable, given human processing constraints.

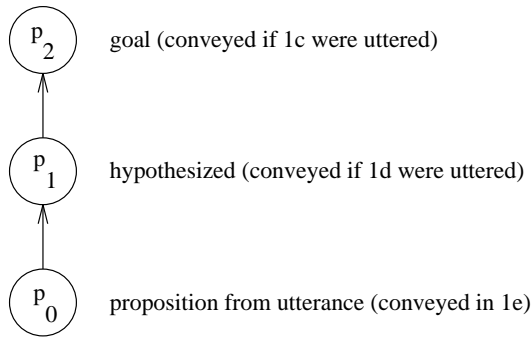

Figure 6: Hypothesis generation tree relating (1e) to (1c)

The set of candidate plans inferred from a response are ranked using two preference criteria.[10] First, as the number of hypothesized propositions in a candidate increases, its plausibility decreases. Second, as the number of non-hypothesized propositions accounted for by the plan increases, its plausibility increases.

To summarize the interpretation algorithm, it is primarily expectation-driven in the sense that the answer interpreter attempts to interpret R's response as an answer generated by some answer discourse plan operator. Whenever the answer interpreter is unable to relate an utterance to the plan which it is currently attempting to recognize, the answer interpreter attempts to find a connection by hypothesis generation. Logical inference plays a supplementary role, namely, in consistency checking (including inferring the plausibility of coherence relations) and in hypothesis generation.

## 4. GENERATION

The inputs to generation consist of 1) the same sets of discourse plan operators and coherence rules used in interpretation, 2) R's beliefs, and 3) the same discourse expectation. The output is a discourse plan for an answer (indirect, if possible). Generation of an indirect reply has two phases: 1) content planning, in which the generator creates a discourse plan for a full answer, and 2) plan pruning, in which the generator determines which parts of the planned full answer do not need to be explicitly stated. For example, given an appropriate set of R's beliefs, our system generates a plan for asserting only the proposition conveyed in (1e) as an answer to (1b).[11]

Content-planning is performed by top-down expansion of an answer discourse plan operator. Note that applicability conditions prevent inappropriate use of an operator, but they do not model a speaker's motivation for providing extra information. Further, a full answer might provide too much information if every satellite whose operator's applicability conditions held were included in a full answer. On the other hand, at the time R is asked the question, R may not yet have the primary goals of a potential satellite. To overcome these limitations, we have incorporated *stimulus conditions* into the discourse plan operators in our model. As mentioned in section 2, stimulus conditions can be thought of as triggers or motivating conditions which give rise to new speaker goals. By analyzing the speaker's possible motivation for providing extra information in the examples in our corpus, we have identified a small set of stimulus conditions which reflect general concerns of accuracy, efficiency, and politeness. In order for a satellite to be included in a full answer, all of its applicability conditions and at least one of its stimulus conditions must hold. (A theorem prover is used to search for an instantiation of the existential variable satisfying the above conditions.)

The output of the content-planning phase, a discourse plan representing a full answer, is the input to the plan-pruning phase. The goal of this phase is to make the response more concise, i.e. to determine which of the planned acts can be omitted while still allowing Q to infer the full plan. To do this, the generator considers each of the acts in the frontier of the full plan tree from right to left (thus ensuring that a satellite is considered before its nucleus). The generator creates trial plans consisting of the original plan minus the nodes pruned so far and minus the current node. Then, the generator simulates Q's interpretation of the trial plan. If Q could infer the full plan (as the most preferred plan), then the current node can be pruned. Note that, even when it is not possible to prune the direct answer, a benefit of this approach is that it generates appropriate extra information with direct answers.

---

[9] Note that, in general, any nodes on the path between $p_0$ and $p_h$, where $p_h$ is the hypothesis returned, will be used as additional hypotheses (later) to connect what was said to $p_h$.

[10] Another possible criterion is whether the actual ordering reflects the default ordering specified in the discourse plan operators. We plan to test the usefulness of this criterion.

[11] The tactical component must choose an appropriate expression to refer to R's car's timing belt, depending on whether (1d) is omitted.

## 5. RELATED RESEARCH

It has been noted [Diller, 1989, Hirschberg, 1985, Lakoff, 1973] that indirect answers *conversationally implicate* [Grice, 1975] direct answers. Recently, philosophers [Thomason, 1990, McCafferty, 1987] have argued for a plan-based approach to conversational implicature. Plan-based computational models have been proposed for similar discourse interpretation problems, e.g. indirect speech acts [Perrault and Allen, 1980, Hinkelman, 1989], but none of these models address the interpretation of indirect answers. Also, our use of coherence relations, both 1) as constraints on the relevance of indirect answers, and 2) in our hypothesis generation algorithm, is unique in plan-based interpretation models.

In addition to RST, a number of theories of text coherence have been proposed [Grimes, 1975, Halliday, 1976, Hobbs, 1979, Polanyi, 1986, Reichman, 1984]. Coherence relations have been used in interpretation [Dahlgren, 1989, Wu and Lytinen, 1990]. However, inference of coherence relations alone is insufficient for interpreting indirect answers, since additional pragmatic knowledge (what we represent as discourse plan operators) and discourse expectations are necessary also. Coherence relations have been used in generation [McKeown, 1985, Hovy, 1988, Moore and Paris, 1988, Horacek, 1992] but none of these models generate indirect answers. Also, our use of stimulus conditions is unique in generation models.

Most previous formal and computational models of conversational implicature [Gazdar, 1979, Green, 1990, Hirschberg, 1985, Lascarides and Asher, 1991] derive implicatures by classical or nonclassical logical inference with one or more licensing rules defining a class of implicatures. Our coherence rules are similar conceptually to the licensing rules in Lascarides et al.'s model of temporal implicature. (However, different coherence relations play a role in indirect answers.) While Lascarides et al. model temporal implicatures as defeasible inferences, such an approach to indirect answers would fail to distinguish what R intends to convey by his response from other default inferences. We claim that R's response in (1), for example, does not warrant the attribution to R of the intention to convey that the rear axle of R's car is made of metal. Hirschberg's model for deriving scalar implicatures addresses only a few of the types of indirect answers that our model does. Furthermore, our discourse-plan-based approach avoids problems faced by licensing-rule-based approaches in handling backward cancellation and multiple-utterance responses [Green and Carberry, 1992].

Also, a potential problem faced by those approaches is scalability, i.e., as licensing rules for handling more types of implicature are added, rule conflicts may arise and tractability may decrease. In contrast, our approach avoids such problems by restricting the use of logical inference.

## 6. CONCLUSION

We have described our implemented computational model for interpreting and generating indirect answers to Yes-No questions. Its main features are 1) a discourse-plan-based approach to implicature, 2) a reversible architecture, 3) a hybrid reasoning model, and 4) use of stimulus conditions for modeling a speaker's motivation for providing appropriate extra information. The model handles a wider range of types of indirect answers than previous computational models. Furthermore, since Yes-No questions and their answers have features in common with other types of *adjacency pairs* [Levinson, 1983], we expect that this approach can be extended to them as well. Finally, a discourse-plan-based approach to implicature has significant advantages over a licensing-rule-based approach. In the future, we would like to integrate our interpretation and generation components with a dialogue system and investigate other factors in generating indirect answers (e.g. multiple goals, stylistic concerns).

## References

[Allen, 1979] James F. Allen. *A Plan-Based Approach to Speech Act Recognition*. PhD thesis, University of Toronto, Toronto, Ontario, Canada, 1979.

[American Express transcripts, 1992] American Express tapes. Transcripts of audiotape conversations made at SRI International, Menlo Park, California. Prepared by Jacqueline Kowto under the direction of Patti Price.

[Brown and Levinson, 1978] Penelope Brown and Stephen Levinson. Universals in language usage: Politeness phenomena. In Esther N. Goody, editor, *Questions and politeness: Strategies in social interaction*, pages 56–289. Cambridge University Press, Cambridge, 1978.

[Carberry, 1990] Sandra Carberry. *Plan Recognition in Natural Language Dialogue*. MIT Press, Cambridge, Massachusetts, 1990.

[Clark and Marshall, 1981] H. Clark and C. Marshall. Definite reference and mutual knowledge. In A. K. Joshi, B. Webber, and I. Sag, editors, *Elements of discourse understanding*. Cambridge University Press, Cambridge, 1981.

[Dahlgren, 1989] Kathleen Dahlgren. Coherence relation assignment. In *Proceedings of the Annual Meeting of the Cognitive Science Society*, pages 588–596, 1989.

[Diller, 1989] Anne-Marie Diller. *La pragmatique des questions et des réponses*. In Tübinger Beiträge zur Linguistik 243. Gunter Narr Verlag, Tübingen, 1989.

[Gazdar, 1979] G. Gazdar. *Pragmatics: Implicature, Presupposition, and Logical Form*. Academic Press, New York, 1979.

[Green, 1990] Nancy L. Green. Normal state implicature. In *Proceedings of the 28th Annual Meeting of the Association for Computational Linguistics*, pages 89–96, 1990.

[Green, in preparation] Nancy L. Green. *A Computational Model for Interpreting and Generating Indirect Answers*. PhD thesis, University of Delaware, in preparation.

[Green and Carberry, 1992] Nancy L. Green and Sandra Carberry. Conversational implicatures in indirect replies. In *Proceedings of the 30th Annual Meeting of the Association for Computational Linguistics*, pages 64–71, 1992.

[Grice, 1975] H. Paul Grice. Logic and conversation. In P. Cole and J. L. Morgan, editors, *Syntax and Semantics III: Speech Acts*, pages 41–58, New York, 1975. Academic Press.

[Grimes, 1975] J. E. Grimes. *The Thread of Discourse*. Mouton, The Hague, 1975.

[Halliday, 1976] M. Halliday. *Cohesion in English*. Longman, London, 1976.

[Hinkelman, 1989] Elizabeth Ann Hinkelman. *Linguistic and Pragmatic Constraints on Utterance Interpretation*. PhD thesis, University of Rochester, 1989.

[Hirschberg, 1985] Julia Bell Hirschberg. *A Theory of Scalar Implicature*. PhD thesis, University of Pennsylvania, 1985.

[Hobbs, 1979] Jerry R. Hobbs. Coherence and coreference. *Cognitive Science*, 3:67–90, 1979.

[Horacek, 1992] Helmut Horacek. An Integrated View of Text Planning. In R. Dale, E. Hovy, D. Rösner, and O. Stock, editors, *Aspects of Automated Natural Language Generation*, pages 29–44, Berlin, 1992. Springer-Verlag.

[Hovy, 1988] Eduard H. Hovy. Planning coherent multisentential text. In *Proceedings of the 26th Annual Meeting of the Association for Computational Linguistics*, pages 163–169, 1988.

[Lakoff, 1973] Robin Lakoff. Questionable answers and answerable questions. In Braj B. Kachru, Robert B. Lees, Yakov Malkiel, Angelina Pietrangeli, and Sol Saporta, editors, *Papers in Honor of Henry and Renée Kahane*, pages 453–467, Urbana, 1973. University of Illinois Press.

[Lascarides and Asher, 1991] Alex Lascarides and Nicholas Asher. Discourse relations and defeasible knowledge. In *Proceedings of the 29th Annual Meeting of the Association for Computational Linguistics*, pages 55–62, 1991.

[Levinson, 1983] S. Levinson. *Pragmatics*. Cambridge University Press, Cambridge, 1983.

[McCafferty, 1987] Andrew Schaub McCafferty. *Reasoning about Implicature: a Plan-Based Approach*. PhD thesis, University of Pittsburgh, 1987.

[McKeown, 1985] Kathleen R. McKeown. *Text Generation*. Cambridge University Press, 1985.

[Mann and Thompson, 1987] W. C. Mann and S. A. Thompson. Rhetorical structure theory: Toward a functional theory of text organization. *Text*, 8(3):167–182, 1987.

[Moore and Paris, 1988] Johanna D. Moore and Cecile L. Paris. Constructing coherent text using rhetorical relations. In *Proceedings of the 10th Annual Conference of the Cognitive Science Society*, August 1988.

[Perrault and Allen, 1980] R. Perrault and J. Allen. A plan-based analysis of indirect speech acts. *American Journal of Computational Linguistics*, 6(3-4):167–182, 1980.

[Polanyi, 1986] Livia Polanyi. The linguistics discourse model: Towards a formal theory of discourse structure. Technical Report 6409, Bolt Beranek and Newman Laboratories Inc., Cambridge, Massachusetts, 1987.

[Reichman, 1984] Rachel Reichman. Extended person-machine interface. *Artificial Intelligence*, 22:157–218, 1984.

[Stenström, 1984] Anna-Brita Stenström. Questions and responses in english conversation. In Claes Schaar and Jan Svartvik, editors, *Lund Studies in English 68*. CWK Gleerup, Malmö, Sweden, 1984.

[Thomason, 1990] Richmond H. Thomason. Accommodation, meaning, and implicature: Interdisciplinary foundations for pragmatics. In P. Cohen, J. Morgan, and M. Pollack, editors, *Intentions in Communication*. MIT Press, Cambridge, Massachusetts, 1990.

[Wu and Lytinen, 1990] Horng Jyh Wu and Steven Lytinen. Coherence relation reasoning in persuasive discourse. In *Proceedings of the Annual Meeting of the Cognitive Science Society*, pages 503–510, 1990.